# Tolerance Principle and Small Language Model Learning

**Adam E. Friedman, Stevan Harnad, and Rushen Shi**

## 1. Introduction
### 1.1. Tolerance Principle (TP)

Unsupervised learning of a rule (learning from passive exposure with no corrective feedback) from a training set of examples requires the ability to generalize the rule to novel instances not seen in the training set (Huebner et al., 2021). Let us call a rule productive if it is learnable from a training set. A theory that can predict and explain whether a rule will be productive given a particular training set would be important in linguistics and cognitive science for the light it would shed on the process of early language acquisition in humans.

One theory of rule generalization is the Tolerance Principle (TP), derived mathematically by Charles Yang (2016) as a necessary consequence of a rule-ordering algorithm known as the Elsewhere Condition (Anderson, 1969; Kiparsky, 1973). Yang proposes the TP as a cognitive model of processing rules and exceptions. According to the Elsewhere Condition, as applied to the human brain, learning operates in an "exceptions-first, rule-later" fashion. When encountering a new exemplar and needing to decide whether to apply a rule, the brain must first consider every known exception to the rule (to see whether this exemplar is one of the known exceptions) before the general rule can be applied to it. When there are very many exceptions and very few rule-following examples, it is more time-efficient to just memorize each exemplar on a case-by-case basis and not try to learn a rule at all. The TP explores, mathematically, the relationship between the number of exceptions and the number of rule-following examples that allows the brain to "optimize/minimize the time complexity of language use," (Yang, 2016, p. 60).

As described and made most explicit in Yang (2018, 2023), "The TP is first and foremost a theory of learning. It specifies a precise threshold, as a proportion of items in the learner's experience, that a generalization can tolerate as exceptions: $\theta_N = N/lnN$, where $N$ is the cardinality of the item set," (Yang, 2023, p. 2). Yang makes the claim that the TP is applicable to any kind of learning

* Adam E. Friedman, McGill University, adam.friedman@mail.mcgill.ca; Stevan Harnad, Université du Québec à Montréal and McGill University, harnad.stevan@uqam.ca; Rushen Shi, Université du Québec à Montréal, shi.rushen@uqam.ca. This study is supported by the Natural Sciences and Engineering Research Council of Canada through a discovery grant to Rushen Shi. We extend our gratitude to Christian Theriault and Hugues Leduc for research assistance.

where a rule must be generalized despite the possibility of exceptions. It is not explicitly limited to natural language rules.

One key feature of the TP is the hypothesis that rule learning does not occur gradually; it is instead quantal, meaning a rule is either productive or unproductive on a given set. Given a sufficient number of examples, (unsupervised) learners should either be able to generalize a rule, or be completely unable to do so, in which case they can only memorize the examples they were given on a case-by-case basis. This applies to learning rules over an entire set or to learning sub-rules over subsets of a set.

Also essential for understanding the TP is that the set size $N$, as well as the number of permissible exceptions $e \leq \theta_N$, both refer to the frequency of unique item "types" in the training set (e.g., in English past tense, "blinked," "googled," are types), not the frequency of "tokens" (occurrences) of the type. The TP posits that as long as a learner is exposed to enough different item types to allow rule learning to occur at all, the number of repetitions of the same type will not affect the productivity of the rule.

To our knowledge, the TP has not been tested on an unsupervised machine learning model prior to this study.

**1.2. Rule Generalization in Human Infants**

The precise effects of type and token frequency on productivity, as well as the matter of quantalness or gradience in the acquisition of productivity, are not universally agreed upon. Numerous studies suggest differences in the learning mechanisms of children and adults. In an artificial language learning experiment conducted by Schuler et al. (2016), children aged 5 to 7 years showed quantal productivity based only on type frequency, aligning well with the TP, while adults appeared to probability match based only on the token frequency. Contrarily, in another artificial language learning experiment with adults, Jarosz et al. (2025) reported that type and token frequencies were jointly responsible for productivity. Children's tendency to regularize, consistent with quantalness, and adults' tendency to probability match, consistent with gradience, are further reported in the artificial language learning experiments of Hudson Kam & Newport (2005, 2009) and Austin et al. (2022).

In the present study we consider a line of experiments that investigated the ability of human infants to generalize grammatical rules. In particular, Koulaguina & Shi (2013) showed that infants aged as young as 14 months can generalize grammar rules to novel instances from relatively little training (as few as 8 exemplar sentences, repeated four times). Koulaguina & Shi (2019) showed with 14-month-olds that a training set that consisted of 50% rule-following and 50% non-rule-following sentences was insufficient for a word-order shift rule to be generalized, while a training set consisting of 80% rule-following and 20% non-rule-following was sufficient. They also found that it was the type frequency in the example set and not the token frequency that determined whether the word-order shift rule was productive.

Shi & Emond (2023) continued the above paradigm with more rigor, attempting to find a threshold of permissible exceptions beyond which generalizability would be impossible. They found that, for a training set of 16 sentences, 14-month-olds could learn a rule when there were 11 rule-following exemplars and 5 exceptions, but could not learn when the input consisted of 10 rule-following sentences and 6 exceptions, consistent with the prediction of the TP. Specifically, for this training set size (N=16), TP predicts a threshold at 5.77, i.e., rule exemplars making up ~63.9% of the input. Shi and Emond also found that babies performed similarly well in the 68.75% rule-following, 80% rule-following, and 100% rule-following cases. They performed similarly poorly in the 50% case and the 62.5% case. These findings suggested a quantal effect across the TP threshold, lending significant support to the TP.

### 1.3. Motivation

It is difficult to explain how or why 14-month-olds are so remarkably capable of generalizing rules to novel instances; however, computational models are less of a black box than a human brain. When a model uses unsupervised learning to learn a rule from noisy or exception-filled data, is its learning governed by the TP, or something like it? This was the question that motivated our work. If it is possible to show that models can do the same thing human infants can do, examining how they do it might help explain how human infants do it.

The problem of explaining the capacity to learn is also at the forefront of language model research (see Contreras et al., 2023; Jawahar et al., 2019). Whereas there is plenty of research on how LLMs learn when provided with superhuman amounts of data and training, there is very limited research on their capacity to learn with small amounts of unlabeled training data through unsupervised learning.

A few efforts have been made to optimize LLMs to achieve substantial learning from developmentally feasible quantities of training data. In the first BabyLM challenge (Warstadt et al., 2023), language models were optimized to maximize learning with a training data size of 10M words or less. Huebner et al. (2021) developed the BabyBERTa transformer-based language model as a variation of RoBERTa-base (Liu et al., 2019) and pre-trained it on as few as 5M words, simulating the input available to children aged up to six years old. Some of the best performers on the first BabyLM challenge used BabyBERTa (Warstadt et al., 2023).

### 2. Implementation
### 2.1. Task

To test the scope and generality of the TP, we address the following questions: (1) What is the minimal amount of training data that our language model needs to learn a rule? (2) How noisy can this training data be? In other words, what proportion of training data in the training set can be rule-violating yet still leave the rule learnable? What is the relation between this proportion and the size of the dataset? (3) Is our language model's productivity quantal or gradient?

**2.2. Model Architecture**

We implement BabyBERTa (Huebner et al., 2021), whose code is open-source and available on GitHub. BabyBERTa uses the Transformers architecture (Vaswani et al., 2017) and is the result of a fine-tuning of the hyper-parameters of RoBERTa (Liu et al., 2019). In its original implementation, it was proposed as a good model for training on cognitively plausible quantities of training data, and it was used successfully for testing grammatical rule learning after training.

BabyBERTa, in line with RoBERTa and differing from BERT (Devlin et al., 2018), does not do next-sentence prediction. It is instead trained only on the masked language model (MLM) pre-training objective used by BERT. A new random subsample of tokens is selected for masking every epoch.

Unlike RoBERTa-base, BabyBERTa is trained exclusively on single sentences. This means that the prediction of masked tokens takes into account only the rest of the tokens in the same sentence as the masked token. The MLM procedure is a form of self-supervised learning.

Like the original BabyBERTa implementation, our model uses 8 layers, 8 attention heads, 256 hidden units, and an intermediate size of 1024. We use Adam optimizer (Kingma & Ba, 2015) with a learning rate of $1e-4$. Batch size is set to 16. In creating a random subsample of tokens for masking, tokens are selected with a probability of 0.15.

**3. Experiment 1**
**3.1. Training Procedure**

We trained many BabyBERTa models separately and from scratch on custom training datasets that contained varying proportions of rule-following and exception exemplars. These datasets also varied in size, i.e., the overall number of exemplars. Each BabyBERTa model was trained on its own unique training dataset in the form of a text (.txt) file. One of the reasons for using a transformers-based architecture is to be able to train our model on sequential text data.

For our first experiment, we based our rule on the word-order rules in the baby experiments of Koulaguina & Shi (2013, 2019) and Shi & Emond (2023). For each BabyBERTa model we trained, the procedure was as follows: A vocabulary set of 10,000 words was generated randomly and saved as a text (.txt) file; each word had 3 to 13 random alphabetic characters. The vocabulary set was partitioned into two—the first 5,000 words were used to create the training set, and the last 5,000 words were used to create the testing set, so as to avoid overlap between the two. From the first 5,000 words, three words were chosen randomly for each sentence in the training set. Those three words were then followed by the same three words again but in a different order. Rule-following sentences had an *ABC-BAC* word-order shift, and exceptions had an *ABC-ACB* word-order shift.

Across training sets, we manipulated (a) the proportion of rule-following sentences to exception sentences, and (b) the total number of sentences.

When we write a sentence as having the form *ABC-BAC*, its actual representation in the training data has the form [A B C B A C . \n], where 'A,' 'B,' and 'C,' are unique tokens from the vocabulary set and '\n' is a newline character. This setup allows the model to consider each exemplar sentence individually, since BabyBERTa is trained exclusively on single sentences.

(1) a. ZTlnz Qih KQxiZUQ Qih ZTlnz KQxiZUQ . \n
     A    B      C      B     A      C

   b. izMewz gLkh VljC izMewz VljC gLkh . \n
     A     B     C     A     C     B

BabyBERTa uses a sub-word vocabulary using the Python API *tokenizers*. A tokenizer performs the necessary function of allowing a language model to break down its input into units that can be processed.

We trained each BabyBERTa model on its unique training set, and trained its respective tokenizer on both the vocabulary set and the training set. We did not use any pre-trained models trained on external datasets, and we did no fine-tuning.

**3.2. Evaluation Procedure**

The goal of evaluation is to test whether a model has learned the word-order shift rule and can generalize it to novel instances unseen during training. After a full training sequence was complete, we tested the trained models on novel test sets whose format was inspired by the grammar test suites used to evaluate BabyBERTa (Huebner et al., 2021), which themselves were inspired by BLiMP (Benchmark of Linguistic Minimal Pairs, Warstadt et al., 2020). To succeed on our tests, a model must be able to consistently give a higher score to a sentence that follows the trained rule than a similar sentence that follows a pattern never seen in training, even when both sentences are made up of novel words.

Test sentences were generated in pairs. For each pair, three words were chosen randomly from the latter 5000 words in the vocabulary set. One sentence in each pair followed the trained rule (*ABC* followed by *BAC*) and the other sentence had the same words *A*, *B*, and *C*, but following a different order that was never seen in training—ABC-CAB. During evaluation, a trained BabyBERTa model assigns a surprisal score to each sentence in its testing set. Surprisal is mathematically the sum of the cross-entropy errors of all the words in the sentence, and it can be used as a metric for how surprising the sentence is to the model. For each test pair, we say that the model prefers the less surprising sentence—the one with a lower surprisal score.

For each BabyBERTa model we trained, we generated a unique test set. Every test set had 1,000 sentence pairs. We calculated each BabyBERTa model's accuracy as the percentage of test trials in which it preferred the trained sentence in a pair. Notably, there is no backpropagation during the testing procedure; the model's weights are frozen after training. This allows the model's rule learning to be probed without risking additional learning during the test.

### 3.3. Results

For every training and testing set we generated, we first trained a BabyBERTa model on four repetitions (i.e., four "epochs") of every item in the training data before evaluating that model, then trained another BabyBERTa model on ten repetitions ("epochs") of the same data before evaluating it.

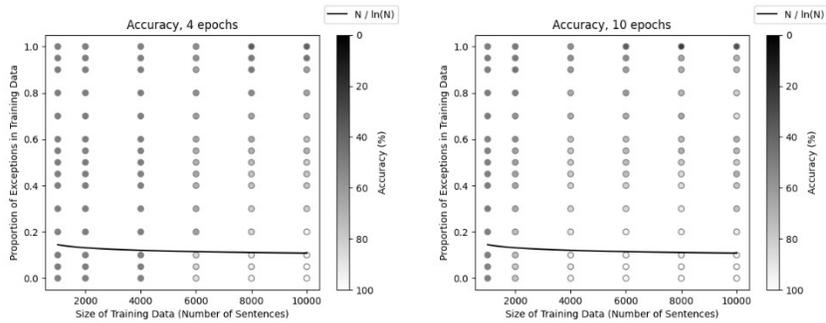

**Figure 1. Model accuracies from Experiment 1, initial trials**

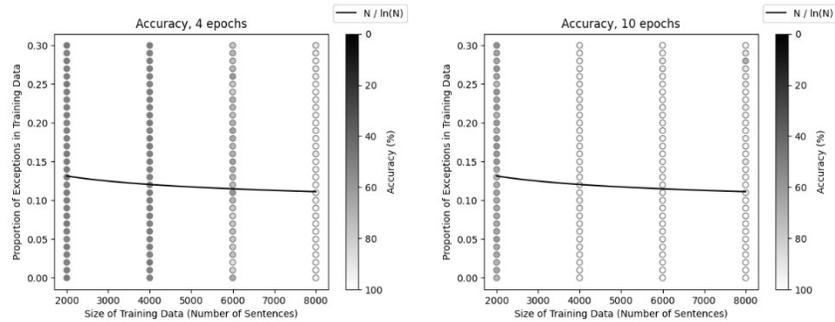

**Figure 2. Model accuracies from Experiment 1, dense trials in range [0, 0.3] of proportions of exceptions in training**

Figures 1 and 2 show the results of training and testing of our models. Each data point is the result of a separate BabyBERTa model; model accuracy during evaluation is represented by how light the point is. We varied training set sizes (x-axis), proportions of exceptions in the training sets (y-axis), and number of epochs of training (graph titles). The threshold of permissible exceptions predicted by the TP is shown by the black curve—the function $N/lnN$, where $N$ is the number of types in the training data. (We plot $1/lnN$ on the above, because the y-axis represents a proportion, not a number.)

The bottom row of data points in each figure shows the accuracies of the models that were trained on fully consistent training input, where all exemplars in their training were rule-following, with no exceptions. Figures 1 and 2 reveal that when the model is exposed to four repetitions of the training data, no rule learning occurs for training set sizes of 4,000 sentences or fewer, even in the ideal case of no exceptions. With ten repetitions, BabyBERTa can learn from fewer examples; no learning is seen when there are 1,000 sentences in the training set, but some learning occurs with 2,000. This finding is consistent with the consensus in the literature that machine learning models see initial improvements in performance after an increase in the number of epochs of training, although it goes against the prediction of the TP which states that the number of repetitions of the same items in training—the token frequency—should not impact a rule's productivity.

The TP predicts that when the number of exceptions does not exceed the threshold ($N/lnN$), there will be full learning, and when the number of exceptions does exceed the threshold, there will be a total failure to learn. Accordingly, TP-like behavior would correspond to successful learning occurring below the black curves in the graphs in Figures 1 & 2 and failing completely above the black curves. However, high model accuracies are consistently seen in regions above the black curves, suggesting by another metric that BabyBERTa's performance is poorly predicted by the TP. Yet, this result does not preclude the possibility that the TP threshold holds some significance in the model's performance. Is there any significant drop in accuracy as proportion of exceptions crosses the threshold? Or is the threshold entirely insignificant to BabyBERTa's learning?

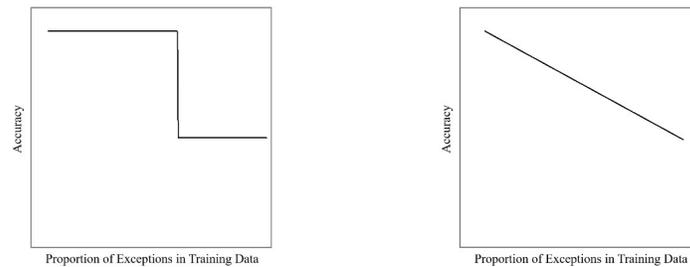

**Figure 3. Ideal quantal graph (left) vs. ideal gradual learning graph (right)**

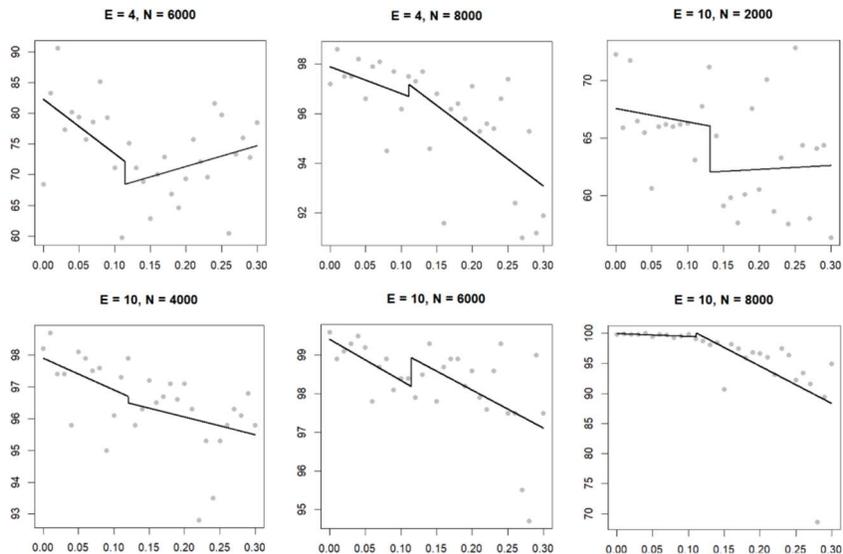
**Figure 4. Linear-regression-based test for quantalness in Experiment 1 data**

Figure 3, left, depicts an all-or-none step function representing the expected change in accuracy as the proportion of exceptions is increased if the model's learning behavior is totally quantal in the ideal sense. Such behavior aligns with the predictions of the TP. Figure 3, right, depicts the expected relationship between model accuracy and the proportion of exceptions in the training data if the model's learning is gradual in the ideal sense, i.e., accuracy gradually decreases as the proportion of exceptions increases.

To test for quantalness, we considered individual columns of data from our dense trials (Figure 2). By considering just one column at a time, we fixed the training dataset size and number of epochs of training. We plotted the effect of increasing the proportion of exceptions in the training data on model accuracy. We then took a linear regression of all of the data to the left of the threshold, where the exceptions were below the tolerance threshold, and another linear regression on the right side of the threshold, where the exceptions were above the threshold. We stitched these regressions together with a vertical bar. Using a t-test, we checked for the significance of the jump from one regression to the next.

Figure 4 shows the results of these statistical tests; we show only the data from columns where some learning occurs, excluding those where model accuracy hovers around 50%. For all six tests shown, the jump in accuracy at the tolerance threshold was not significant. Note that since the TP threshold is a function of the size of the training data, the jump does not always appear in the same place. Statistically, there will always be a jump-like effect at the TP threshold, where we merged the two regression lines with a vertical bar; this is an artifact of the stochastic nature of the data. These jumps are usually not significant. They are also not to scale due to the varying Y-axis scales in Figure 2.

## 4. Experiment 2
### 4.1. Training Procedure

The Tolerance Principle was proposed to apply not only to the unsupervised learning of grammatical rules, but to the unsupervised learning of rules and categories in general. For our second experiment, we trained and tested our BabyBERTa models on a non-grammar rule which was as simple as possible.

As in Experiment 1, we trained our models on text (.txt) files. We trained the models on binary strings of 0's and 1's of length 16. Our binary rule was: the first digit of each string should be '1', such as the example in 2a. Exceptions to the rule were strings that started with '0', such as the example in 2b. This rule is simpler and less grammar-like than the rule used in Experiment 1. It has only one defining feature.

(2) a. 1001011101010100 \n
    Rule-following

   b. 0101110101001010 \n
    Exception

When generating our training data, we only controlled the first digit of the strings. The rest of the digits were randomly generated. Across training sets, we varied (a) the proportion of rule-following strings to exception strings, and (b) the total number of strings in the training sets.

### 4.2. Evaluation Procedure

We generated 1,000 sentence pairs in each test set used for evaluation. The two sentences in each pair were completely identical binary strings except that the first digit of the 16 digits of one of the strings was '1' and in the other it was '0'. After a model was trained, we tested its learning of the rule by calculating surprisal scores for every sentence in the test set. If the model has learned a rule, then it should assign a lower surprisal score to a string that follows the rule than to a nearly identical string that breaks the rule.

Unlike in Experiment 1, we now evaluated the models' preference for the dominant pattern seen in training over a minority pattern also seen in training, as opposed to preference for a pattern seen in training over a pattern never seen in training. This is because our binary paradigm, with just one relevant feature, only allows for two possible patterns. We compute model accuracy as the percentage of sentence pairs for which the model preferred the rule-following pattern.

### 4.3. Results

We found that for this simple rule, BabyBERTa can learn from fewer examples than were necessary in Experiment 1, which means shorter training times. Due to the reduced constraint on computation time, for each training set, we trained multiple BabyBERTa models on that set with random weight initialization. Then, using a test set with novel sentences, we evaluated each of those models on that test set. We then averaged their accuracy scores. This has the effect of reducing noise in our results caused by randomness in the model's initialization, yielding results more representative of the model's performance in general and reducing the influence of outliers.

For every training and testing set, we averaged the scores of three BabyBERTa models trained on four repetitions of the training input and did the same for another three BabyBERTa models trained on ten repetitions of the training input.

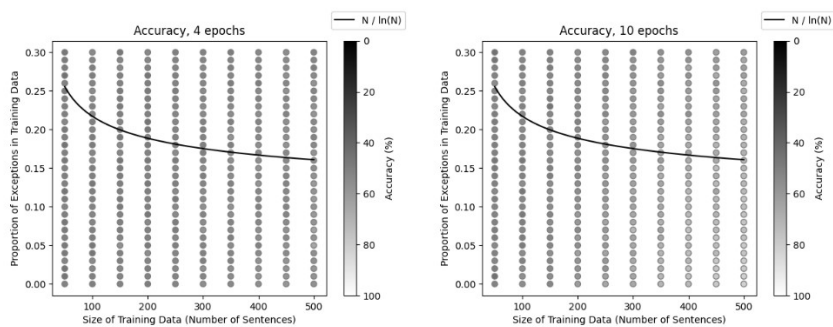

**Figure 5. Averaged model accuracies from Experiment 2**

Figure 5 shows the results of training and testing our models. Each data point is the average of three BabyBERTa models' scores on evaluation. For each of these three models, as described above, the training dataset was the same, and the novel exemplars used for evaluation (unseen during training) were the same. Training took place for the same number of epochs. The only difference between these three models is the initial model weights.

Average accuracy on evaluation is represented by how light a point is. We varied training set sizes (x-axis), proportions of exceptions in the training sets (y-axis), and number of epochs of training (graph titles). The TP threshold is shown by the black curve.

In the bottom-most row of data points of Figure 5, left, which are the accuracies of models trained on fully rule-following training input for four repetitions, performance increases to become better than chance after training on sets with as few as 400 exemplars. Increasing the number of repetitions to ten (Figure 5, right) allows the models to reach better-than-chance performance with as few as 200 exemplars in training, with high accuracies in the 300-to-500 range.

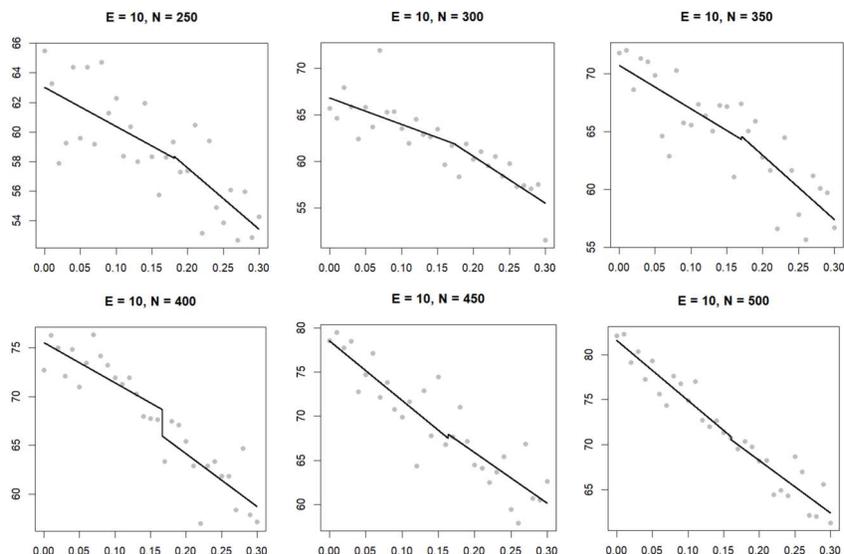
**Figure 6. Linear-regression-based test for quantalness in Experiment 2 data**

Figure 6 shows the linear-regression-based analyses for the six rightmost columns of data from Figure 5, right, which are the regions where our models achieved the highest performance in Experiment 2. In general, we see no TP-like quantal effect at the TP threshold; there were no statistically significant downward jumps at the TP threshold.

In regions of the data where learning occurred, we instead tend to see a gradient decrease in model accuracy as the proportion of exceptions in the training set increases. The trend appears highly similar to Figure 3, right, which is an idealized depiction of gradual learning.

## 5. Discussion

In Experiment 1, using a grammar-like rule, we examined BabyBERTa's preference for a pattern seen during training over a similar pattern unseen during training while varying the proportion of the training input that followed the trained pattern. In Experiment 2, using a simple, non-grammatical rule, we examined BabyBERTa's preference for one pattern seen during training over another pattern also seen during training while varying the ratio of one pattern to the other in the input. In both experiments, we evaluated BabyBERTa's performance at the task of learning a target pattern from training datasets of varying sizes. We also compared the results of training the model for four repetitions of the training input with the results of training for ten repetitions.

Experiment 1 revealed that BabyBERTa requires thousands of exemplar sentences to learn a word-order movement rule that human infants can learn from under ten exemplars, as prior work by Koulaguina & Shi (2013, 2019) and Shi &

Emond (2023) demonstrated. This being the case despite BabyBERTa's supposed optimality for training on cognitively plausible amounts of training data speaks to the data-hungry nature of language models in general. Experiment 2 revealed that for a simpler rule, an order of magnitude less data was required for learning to occur—a few hundred exemplars was enough—suggesting that the complexity of a rule has an influence on the number of instances of the rule needed for BabyBERTa to learn it in unsupervised fashion.

The findings of both Experiments 1 & 2 demonstrated a clear effect of the token frequency on BabyBERTa's rule learning: training for more epochs over the same data increased model accuracy. Experiment 1 especially exemplified the key role of token frequency. Figures 1 & 2 show that there are training dataset sizes (e.g., 4,000 exemplar sentences) for which four repetitions of the training data lead to no learning from BabyBERTa while ten repetitions lead to substantial levels of learning. Further, BabyBERTa achieved similar scores when trained on datasets of 4,000 sentences for ten repetitions as when trained on 10,000 sentences for four repetitions; in token frequency, these two situations are equivalent. This behavior diverges highly from the TP (Yang 2016), which predicts that it is the type frequency in the training input, not the token frequency, that determines productivity—although our findings do not explicitly rule out the possibility that there is still some role played by type frequency.

We analyzed the effect on model accuracy of increasing the proportion of exception exemplars in the training input. We found that, unlike with infants, (a) learning was not quantal; accuracy decreased gradiently as the proportion of exceptions in the training input was increased, (b) the tolerance threshold predicted by TP held no significance to BabyBERTa and was a poor predictor of productivity, and, (c) the prior two points held true both when evaluating preference for a seen pattern over an unseen pattern (as in Experiment 1) and when evaluating preference for a seen majority pattern over a seen minority pattern (as in Experiment 2).

In short, BabyBERTa's behavior does not align with the regularization and quantal productivity observed in human children (Koulaguina & Shi, 2013, 2019; Shi & Emond, 2023; Hudson Kam & Newport, 2005, 2009; Schuler et al., 2016; Austin et al., 2022).

To conclude: for this machine learning architecture, datasets of a few hundred to a few thousand exemplars are necessary for rule learning to occur, and increasing the complexity of the rule causes the model to require an increased number of exemplars to learn. Learning appears to follow a gradient: as the proportion of exception types in the training input increases, there is a gradual, not an all-or-none, decrease in accuracy. As overall token frequency increases accuracy increases; training for more epochs over the same training input increases accuracy. The threshold predicted by the TP seems to have no significant bearing on the language model's learning. Together, these findings suggest that BabyBERTa functions with a different learning mechanism than human infants.

# References

Anderson, Stephen R. (1969). West Scandinavian vowel systems and the ordering of phonological rules. PhD thesis, MIT.

Austin, Alison C., Schuler, Kathryn D., Furlong, Sarah, & Newport, Elissa L. (2022). Learning a language from inconsistent input: Regularization in child and adult learners. *Language Learning and Development*, *18*(3), 249-277.

Contreras Kallens, Pablo, Kristensen-McLachlan, Ross Deans, & Christiansen, Morten H. (2023). Large language models demonstrate the potential of statistical learning in language. *Cognitive Science, 47*(3), e13256.

Devlin, Jacob, Chang, Ming-Wei, Lee, Kenton, & Toutanova, Kristina (2019, June). Bert: Pre-training of deep bidirectional transformers for language understanding. In *Proceedings of the 2019 conference of the North American chapter of the association for computational linguistics: human language technologies, volume 1 (long and short papers)* (pp. 4171-4186).

Hudson Kam, Carla L., & Newport, Elissa L. (2005). Regularizing unpredictable variation: The roles of adult and child learners in language formation and change. *Language learning and development*, *1*(2), 151-195.

Hudson Kam, Carla L., & Newport, Elissa L. (2009). Getting it right by getting it wrong: When learners change languages. *Cognitive psychology*, *59*(1), 30-66.

Huebner, Philip A., Sulem, Elior, Cynthia, Fisher, & Roth, Dan (2021, November). BabyBERTa: Learning more grammar with small-scale child-directed language. In *Proceedings of the 25th conference on computational natural language learning* (pp. 624-646).

Jarosz, Gaja, Hughes, Cerys, Lamont, Andrew, Prickett, Brandon, Baird, Maggie, Kim, Seoyoung, & Nelson, Max (2025). Type and token frequency jointly drive learning of morphology. *Journal of Memory and Language*, 104666.

Jawahar, Ganesh, Sagot, Benoît, & Seddah, Djamé (2019). What does BERT learn about the structure of language? In *ACL 2019-57th Annual Meeting of the Association for Computational Linguistics*.

Kingma, Diederik P. & Ba, Jimmy (2015, May). Adam: A method for stochastic optimization. In *International conference on learning representations (ICLR)* (Vol. 5, No. 6).

Kiparsky, Paul (1973). Elsewhere in phonology. In Anderson, Stephen R. and Kiparsky, Paul, editors, *A festschrift for Morris Halle*, pages 93–106. Holt, Rinehart and Winston, New York.

Koulaguina, Elena, & Shi, Rushen (2013). Abstract rule learning in 11-and 14-month-old infants. *Journal of psycholinguistic research*, *42*, 71-80.

Koulaguina, Elena, & Shi, Rushen (2019). Rule generalization from inconsistent input in early infancy. *Language Acquisition*, *26*(4), 416-435.

Liu, Yinhan, Ott, Myle, Goyal, Naman, Du, Jingfei, Joshi, Mandar, Chen, Danqi, Levy, Omer, Lewis, Mike, Zettlemoyer, Luke, & Stoyanov, Veselin (2019). Roberta: A robustly optimized bert pretraining approach. *arXiv preprint arXiv:1907.11692*.

MacWhinney, Brian (2000). *The CHILDES project: The database* (Vol. 2). Psychology Press.

Schuler, Katherine D., Yang, Charles, & Newport, Elissa L. (2016). Testing the Tolerance Principle: Children form productive rules when it is more computationally efficient to do so. In *Proceedings of the Annual Meeting of the Cognitive Science Society* (Vol. 38).

Shi, Rushen, & Emond, Emeryse (2023). The threshold of rule productivity in infants. *Frontiers in Psychology*, *14*, 1251124.

Vaswani, Ashish, Shazeer, Noam, Parmar, Niki, Uszkoreit, Jakob, Jones, Llion, Gomez, Aidan N., Kaiser, Lukasz, & Polosukhin, Illia. (2017). Attention is all you need. *Advances in neural information processing systems*, *30*.

Warstadt, Alex, Mueller, Aaron, Choshen, Leshem, Wilcox, Ethan, Zhuang, Chengxu, Ciro, Juan, Mosquera, Rafael, Paranjape, Bhargavi, Williams, Adina, Linzen, Tal, & Cotterell, Ryan (2023). Findings of the BabyLM Challenge: Sample-efficient pretraining on developmentally plausible corpora. In *Proceedings of the BabyLM Challenge at the 27th Conference on Computational Natural Language Learning*.

Warstadt, Alex, Parrish, Alicia, Liu, Haokun, Mohananey, Anhad, Peng, Wei, Wang, Sheng-Fu, & Bowman, Samuel R. (2020). BLiMP: The Benchmark of Linguistic Minimal Pairs for English. *Transactions of the Association for Computational Linguistics*, 8:377–392.

Yang, Charles (2016). *The price of linguistic productivity.* Cambridge, MA: The MIT Press.

Yang, Charles (2018). *A user's guide to the tolerance principle*. LingBuzz, https://ling.auf.net/lingbuzz/004146

Yang, Charles (2023). *A user's defense of the tolerance principle*. University of Pennsylvania, https://www.ling.upenn.edu/courses/ling5700/Yang2023enger.pdf